\documentclass[10pt,twocolumn,letterpaper]{article}

\usepackage[pagenumbers]{cvpr}              

\usepackage{natbib}

\usepackage[accsupp]{axessibility}
\usepackage{algorithm}
\usepackage{algorithmic}
\usepackage{amsmath}
\usepackage{makecell}
\usepackage{multirow}
\usepackage{subcaption}
\usepackage[table]{xcolor}
\usepackage{enumitem}
\setlist[itemize]{
  label=$\bullet$,
  leftmargin=1em,
  labelsep=0.3em,
  itemsep=0pt,
  parsep=0pt,
  topsep=0pt
}

\definecolor{cvprblue}{rgb}{0.21,0.49,0.74}
\usepackage[pagebackref,breaklinks,colorlinks,allcolors=cvprblue]{hyperref}

\def\paperID{22310} 
\def\confName{CVPR}
\def\confYear{2026}

\title{ElasticFormer: Detecting Objects in HRW Shots via Elastic Computing Vision Transformer}

\author{
    \textbf{Wenxi Li}\textsuperscript{\ddag}\thanks{Equal contribution},
    \textbf{Jingchen Huang}\textsuperscript{\S,\P}\footnotemark[1],
    \textbf{Chenyang Lyu}\textsuperscript{$\|$},
    \textbf{Moran Liu}\textsuperscript{\S,\P,$\#$},\\
    \textbf{Haozhe Lin}\textsuperscript{\S},
    \textbf{Guiguang Ding}\textsuperscript{\S,$\star$}\thanks{Corresponding author},
    \textbf{Yuchen Guo}\textsuperscript{\S}\footnotemark[2]\\
    \textsuperscript{\S}Beijing National Research Center for Information Science and Technology, \\Tsinghua University, Beijing, China\\
    \textsuperscript{\P}Department of Automation, Tsinghua University, Beijing, China\\
    \textsuperscript{\ddag}KLATASDS-MOE, School of Statistics, East China Normal University, Shanghai, China\\
    \textsuperscript{$\star$}School of Software, Tsinghua University, Beijing, China\\
    \textsuperscript{$\#$}Zhuoxi Lab, Hangzhou, China\\
    \textsuperscript{$\|$}AI Business, Alibaba International Digital Commerce, Hangzhou, China\\
    wxli@sfs.ecnu.edu.cn, huang-jc23@mails.tsinghua.edu.cn, \{linhz, dinggg\}@tsinghua.edu.cn,\\
    liumoreran@gmail.com, lyuchenyang.dcu@gmail.com, yuchen.w.guo@gmail.com
}

\begin{document}
\maketitle

\begin{abstract}
    Recent advances in gigapixel-level imaging have brought High-Resolution Wide shots  to the forefront of research. However, these images present significant challenges: extreme sparsity of foreground, gigapixel-level resolutions and diverse target counts. This makes traditional close-up detectors inaccurate and slow as they are overwhelmed by the background. Although previous research has explored sparse backbones, their fixed sparsity patterns lack the adaptability required to handle diverse target numbers.
To address this, we introduce ElasticFormer, a sparse backbone that dynamically allocates computational resources based on foreground proportion. After scoring windows based on variance, proposed ElasticSelector module will predict the foreground proportion for top-k selection. The mechanism guides the model to select target-containing windows, scaling resources in areas where objects are clustered.
We introduce a novel loss function combined with the 3-phase training strategy for ElasticSelector, allowing it to function properly when bounding box annotations are missing. A WSOD study is carried on PASCAL VOC 2007 to evaluate its extensibility. Further, ElasticNet is created to verify its backbone-agnostic nature. In experiments on the PANDA gigapixel benchmark, ElasticFormer reduces backbone FLOPs by 80\% while achieving a significant improvement in AP$_{50}$ when compared to fixed-ratio sparse methods. 
\end{abstract}

\section{Introduction}
Object detection is a core task in computer vision that involves both locating and identifying objects of interest in an image. It typically produces, for each detected instance, a category label and a spatial extent (usually an axis-aligned bounding box). Models such as Fast R-CNN \cite{girshick2015fast} have demonstrated strong performance on MS COCO \cite{lin2014microsoft}. However, with the development of nondestructive high-resolution imaging systems, High-Resolution Wide (HRW) shots detection has drawn increasing attention \cite{chen2022towards,fan2022speed}, which refers to object detection in imagery characterized by extremely high spatial resolution and a wide field of view, such as images captured by gigapixel-scale drone \cite{svanstrom2021real}, satellite \cite{DOTA, ding2021object, ding2018learning}, or panoramic camera \cite{360+x}.

\begin{figure}
    \centering
    \includegraphics[width=1\linewidth]{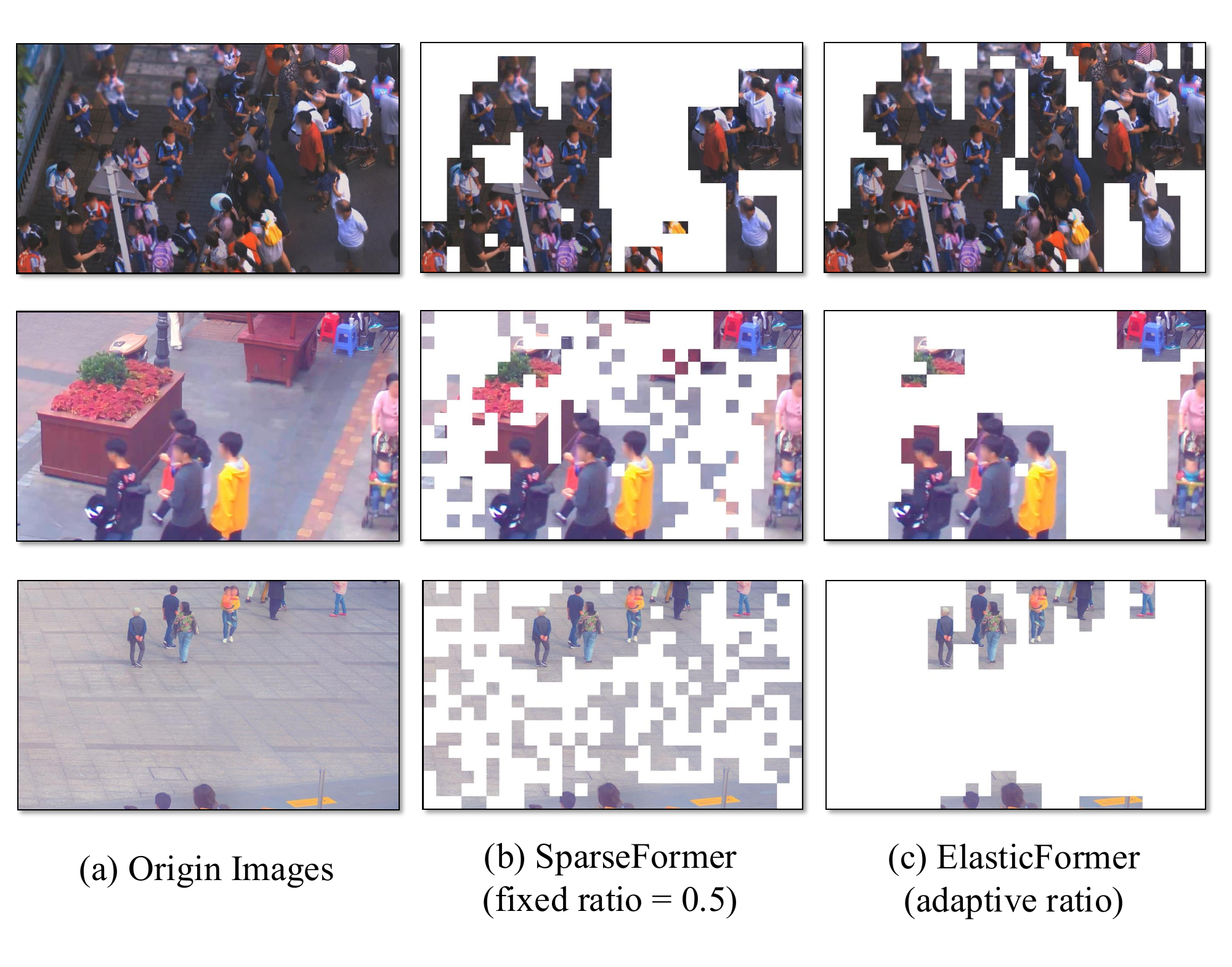}
    \caption{\textbf{Comparison of inference region selection between different backbones.} (b) SparseFormer selects a fixed ratio of regions for processing, regardless of input variability. (c) ElasticFormer adaptively selects a dynamic ratio of informative regions based on the input image content, enabling more flexible and efficient inference.}
    \label{fig:1}
\end{figure}

Firstly, in HRW scenarios, foreground density is much lower than that in conventional detection images. While typical datasets have foregrounds covering over 50\% of the image, HRW images often contain less than 15\% \cite{SparseFormer}, with large background regions such as sky or vegetation. Given such extreme sparsity, cutting-edge detectors, including DINO \cite{DINO}, Deformable DETR \cite{DETR}, RMT \cite{RMT}, among others, still struggle with overwhelming background.

Secondly, limited computational resources pose another challenge. HRW detection scenarios (e.g., UAV-based inspection \cite{VisDrone} and medical detection \cite{VL-Health}) often require real-time responsiveness. As resolution scales from millions of pixels in COCO \cite{lin2014microsoft} to billions in PANDA \cite{PANDA}, the cost of conventional detectors grows polynomially with input size, leading to inference up to two orders of magnitude slower.

Thirdly, HRW scenarios show extreme variation in target counts, ranging from a few to dozens, as shown in \Cref{fig:1}(a). Thus models must remain effective across both sparse and dense cases. However, existing approaches such as SaccadeDet \cite{SaccadeDet} and SparseFormer \cite{SparseFormer} overlook this issue, as they rely on fixed sparsity rates that fail to adapt to fluctuations in target quantity. Consequently, their effectiveness is significantly reduced when confronted with highly variable target distributions, illustrated in \Cref{fig:1}(b).

Inspired by human cognitive efficiency and recent advances \cite{r16,Sparse-RCNN,Adaptformer,CVANet,r20}, we propose ElasticFormer, a detection framework that introduces elasticity into backbone design. Unlike fixed-ratio methods that waste computation in HRW scenes with highly variable foreground proportion, ElasticFormer dynamically allocates resources by adaptively selecting image patches according to the predicted foreground proportion. A lightweight module, ElasticSelector, guides patch selection independently at each stage in backbone, enabling fine-grained perception while scaling cost with foreground proportion rather than overall image size. It is noteworthy that ElasticSelector is backbone-agnostic, motivating the creation of ElasticNet for verification. Experiments show that ElasticFormer reduces computation by up to 80\% while improving accuracy compared with SparseFormer \cite{SparseFormer}, ElasticNet also achieves a 87\% computational saving compared with ResNet-50 \cite{he2016deep}.

A further challenge in weakly supervised object detection (WSOD) is the absence of bounding-box annotations, making foreground proportion hard to compute. To address this, we design a new loss function for ElasticSelector, which uses either ground truth or pseudo labels as supervision. Validation on the PASCAL VOC 2007 dataset \cite{everingham2010pascal} demonstrates that this extension reduces FLOPs by at least 50\% while improving both mAP and CorLoc.

To summarize, our key contributions are: 

\begin{itemize}
\item We propose ElasticFormer, a novel sparse Vision Transformer-based backbone that not only addresses the extremely sparse target distribution and redundant background in HRW images, but also dynamically adapts to varying numbers of targets across diverse scenes.

\item We further introduce a lightweight component, ElasticSelector, which dynamically predicts the foreground proportion during the forward inference process based on intermediate features, providing efficient guidance with negligible additional computational overhead.

\item We design a label-free loss function that minimizes the gap between the predicted and actual foreground proportions, enabling self-supervised optimization of ElasticSelector without the need for extra annotations.

\item We validate the proposed backbone on PANDA \cite{PANDA} and PASCAL VOC 2007 \cite{everingham2010pascal} datasets, and further extend our backbone-agnostic design to CNN-based architectures, consistently demonstrating its effectiveness and generality across model families.
\end{itemize}

\section{Related Work}
\begin{figure*}[t]
    \centering
    \includegraphics[width=1\textwidth]{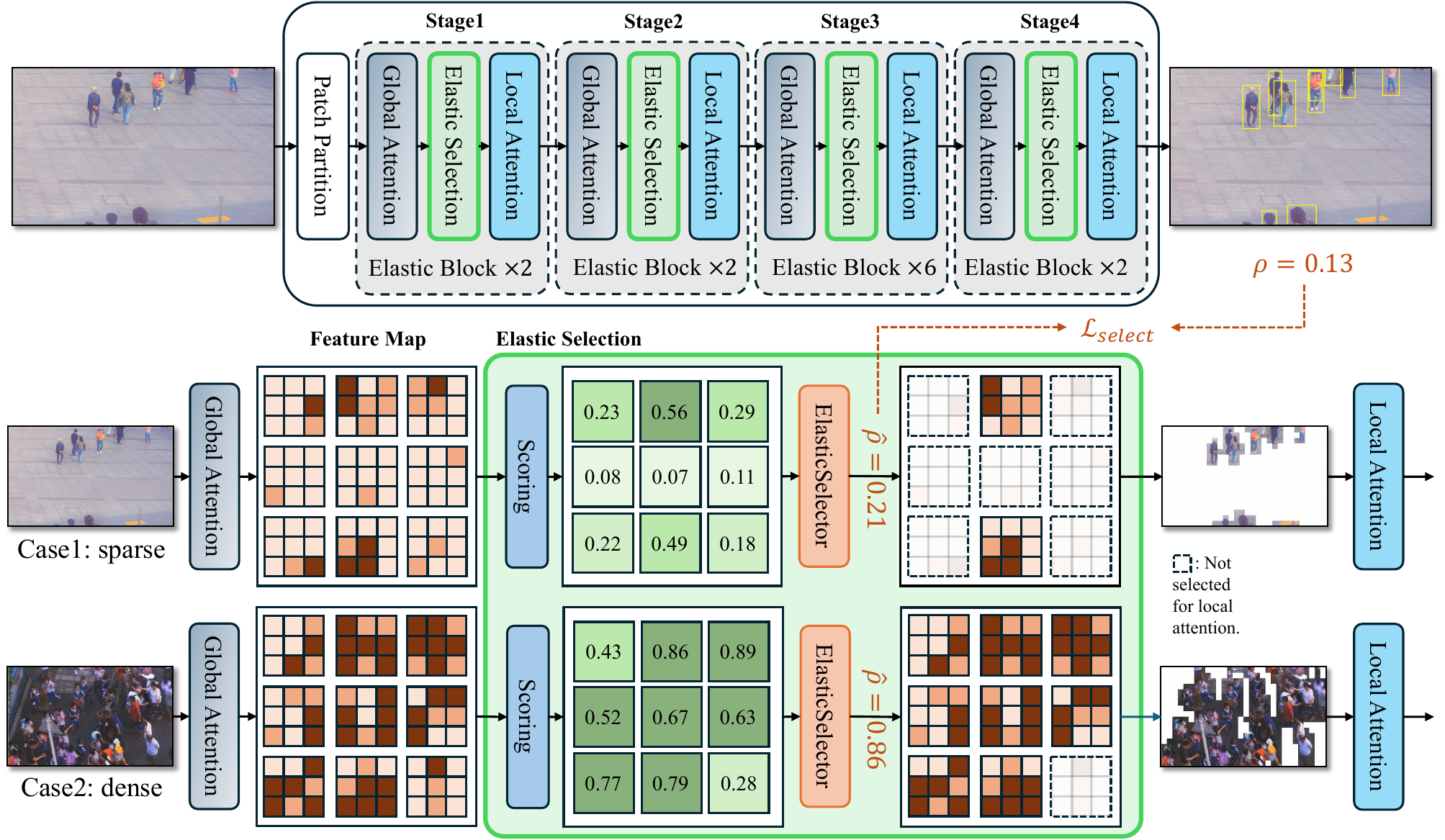}
    \caption{\textbf{Architecture Overview of ElasticFormer.} The green box represents the ElasticBlock,in which the process of ``Scoring" is listed in \cref{eq:1}. Example scoring results of 9 windows is listed here, where the higher score corresponds to the darker color, indicating larger variance and a greater likelihood of containing targets. ElasticSelector selects windows according to their scores and the predicted $\hat{\rho}$, shown in \cref{eq:5}. Here the sparse case select 2 windows out of 9 while the dense case select 8 windows as $\hat{\rho}$ differs. ElasticSelector is trained via $\mathcal{L}_{select}$, in which $\rho$ represents the real foreground proportion.}
    \label{fig:2}
\end{figure*}

\textbf{Object Detection in High-Resolution Wide Shots.}
HRW detection addresses the critical challenge of locating sparse objects in gigapixel-scale imagery, where target coverage is often below 15\% and background is highly diverse \cite{PANDA}, different from ImageNet \cite{Imagenet}. While general-purpose detectors like Faster R-CNN \cite{girshick2015fast} and Deformable Convolutional Networks \cite{DeformableCN} achieve strong performance on dense scenes, their cost scales prohibitively with input resolution, causing severe latency. Shufflenet \cite{Shufflenet} reduces complexity via grouped convolutions, yet lacks adaptive computation. DynamicDet \cite{Dynamicdet} routes images by predicted difficulty, but its binary partitioning is ill-suited to HRW shots and maintaining dual branches is cumbersome. While DGE \cite{DGE} has been applied to enhance vision transformer, it lacks the flexibility required to effectively handle HRW scenarios.

\noindent\textbf{Vision Transformers for Detection.}
Transformers have revolutionized NLP, and Vision Transformers (ViTs) extend this success to detection by modeling long-range dependencies. However, conventional ViTs process all image tokens equally, incurring quadratic complexity with input size. To reduce cost, methods such as window-based attention \cite{Swin,r26}, multi-layer self-attention \cite{PyramidViT}, and low-rank projection \cite{Low-rank} have been proposed. Yet these approaches still face a trade-off between efficiency and accuracy, limiting their effectiveness in HRW detection.

\noindent\textbf{Sparse Computation Methods.}
Sparse detection paradigms reduce redundancy by focusing on informative regions. Dynamic token pruning \cite{Dynamicvit} adjust computation per instance but remains limited by rigid backbones. Saliency-guided approaches like SaccadeDet \cite{SaccadeDet} use auxiliary networks to predict candidate regions but introduce cascaded latency and struggle with extreme scale variations in HRW scenes. Fixed-ratio sparsification, exemplified by SparseFormer \cite{SparseFormer}, retain a preset fraction of tokens, which wastes resources in sparse regions and under-allocates in dense ones. Critically, no existing method dynamically modulates token utilization across backbone stages based on spatially varying target density—a gap our ElasticSelector and ElasticFormer explicitly address.

\section{Method}


\subsection{Overview of ElasticFormer}


In the main network, input image is partitioned into non-overlapping patches of size ($\frac{H}{4}\times \frac{W}{4} \times 48$) where $H$ and $W$ denote the original size of input. Through linear embedding layer, the number of channel is projected into $C$ in the first stage. In subsequent stages, the width and height of the feature maps are progressively halved while channel doubled by merging patches, reshaping the size into ($\frac{H}{8}\times \frac{W}{8} \times 2C$) for global attention. Feature maps from global attention will be partitioned into winodws of size $w\times w$ for local attention, \Cref{fig:2} demonstrates a example of $w=3$, partitioning the feature map into 9 windows. Within each stage, the global attention and local attention modules appear in paired blocks, repeated $d_i$ times. We refer to this repetition count as the depth of that stage, which is set to $[2, 2, 6, 2]$ in ElasticFormer.

In \Cref{fig:2}, we illustrate the structure of an Elastic Selection module in Stage 2 as an example. The feature map is divided into nine windows, each evaluated by the scoring module to obtain a score. Leveraging a variance-based mechanism, the scoring module assigns higher scores to windows containing richer foreground content. For visualization, windows are shaded with different intensities according to their scores: darker colors indicate higher scores and thus more foreground. Two extreme cases of foreground distribution are shown. In the sparse case, only the upper windows receive high scores, while in the dense case, most windows achieve high scores since they all contain substantial foreground regions.

After scoring, ElasticSelector predicts the overall foreground proportion of the image, which is used as the sparsity rate to guide a top‑k selection of windows according to their scores. Discarded windows are marked with black hatching, while the selected ones retain their color and are forwarded to local attention for further foreground feature extraction. The intermediate visualization clearly illustrates how window selection adapts under different foreground proportions, highlighting both sparse and dense cases within the image. Here the proposed $\mathcal{L}_{\text{select}}$ takes the GT foreground ratio $\rho$ as a supervisory signal, guiding ElasticSelector to produce accurate predictions $\hat{\rho}$.

\subsection{Elastic Selection}

\textbf{Definition of foreground proportion.}
To establish a transparent and interpretable basis for adaptive computation allocation within ElasticFormer, we choose the foreground proportion $\rho$ as the guidance. This ratio provides a direct geometric measure of the spatial complexity of an image by quantifying the overall area of all objects. First, we define $\rho$ as:
\begin{equation}
    \label{eq:1}
    \rho = \alpha \cdot \frac{\cup_i \text{BBox}_i}{H \cdot W}.
\end{equation}

Here $\text{BBox}_i$ represent the bounding box of the $i^{th}$ object while $H$ and $W$ denoting the size of complete image. $\cup_i$ means summing up the area of all BBoxes where the overlap area will be counted only once. Since the backbone may not require all regions within the bounding box for effective feature extraction, we introduce a scaling factor $\alpha\in(0,1)$ for $\rho$, encouraging the selection of more critical windows. We fix $\alpha = 0.8$ through experiments. 

\noindent\textbf{Variance-based token scoring.}
Identifying windows most likely to contain targets presents a critical challenge, constituting the key to enabling sparse computation in local attention. Observing that background regions (e.g., sky, plaza and roads) typically exhibit more uniform pixel distributions (i.e., lower variance) than object regions, we score each window as below, where the residual effectively captures the variance.

For the input feature map $\mathbf{F}$ in shape $(B\times H\times W\times C)$, we compute its average pooled result in shape $(B\times \frac{H}{w}\times \frac{W}{w}\times C)$ and interpolate it back. The residual $\mathbf{R}$ is partitioned into $\mathbf{R}_k$ by window shape.
We use $\mathbf{R}_k$ to calculate the score $s_k$ for $k^{th}$ window, then normalize it into $\hat{s}_k$:
\begin{equation}
    \begin{aligned}
        \mathbf{R} &= \mathbf{F} - \text{Interpolate}(\text{AvgPool}_{w \times w}(\mathbf{F}), w) \\
        s_k &= f_{\text{score}}(\text{Flatten}(\mathbf{R}_k)) \quad \forall k \in \{1, \ldots, N_w\} \\
        \hat{s}_k &= \frac{\exp(s_k)}{\sum_{j=1}^{N_w} \exp(s_j)} \\
    \end{aligned}
    \label{eq:4}
\end{equation}
Here $N_w$ is the total number of windows. Through this computational approach, we construct an ordered sequence of windows whose scores are positively correlated with their variances, prepared for sparse selection. We take an example ($N_w=9$) in \Cref{fig:2} to illustrate the score results.

\noindent\textbf{Top-k selection.}
For self-attention module in backbone, limited tokens are needed as background regions do not contain information about objects. We use $\hat{\rho}$ generated by ElasticSelector to select the N highest-scoring windows:
\begin{equation}
    N = \lceil \hat{\rho} \cdot N_w \rceil,
    \label{eq:5}
\end{equation}
where $N_w$ is the total number of windows.

These sequentially selected windows correspond to the regions in the image where target objects are most likely to appear. Taking \Cref{fig:2} as an example, in Case1:sparse, $N=\lceil 0.21\cdot 9 \rceil=2$, thus top-2 windows with score of 0.56 and 0.49 are selected for ``Local Attention".In the following, we detail how ElasticSelector predicts $\hat{\rho}$ that determines the number of windows to be retained.

\subsection{ElasticSelector}

\noindent\textbf{Predicting foreground proportion.}
In order to obtain such a feature of the overall image, $\rho$, we used feature map for extraction through a MLP. Because the MLP’s input dimensions are fixed and restricted to one-dimensional vectors, we apply a size-normalization operation to the feature maps:
\begin{equation}
    \begin{aligned}
        \mathbf{F}_{\text{pooled}} &= \text{AdpPool}(\mathbf{F}), \\
        \mathbf{F}_{\text{flat}} &= \text{Flatten}(\mathbf{F}_{\text{pooled}}).
    \end{aligned}
\end{equation}
The adaptive average pooling (AdpPool) downsamples the input feature map $\mathbf{F}$ (derived from the backbone) to a predetermined, fixed spatial resolution regardless of the original size of input images. Here the size for 4 stages is $[32^2,\ 16^2,\ 8^2,\ 4^2]$. The subsequent Flatten layer then serializes the two-dimensional output $\mathbf{F}_{\text{pooled}}$ into a one-dimensional feature vector $\mathbf{F}_{\text{flat}}$. The dimensionality of this vector is inherently determined by the output size chosen for the adaptive pooling and the number of feature channels $C$, and is strategically configured based on the specific stage within the ElasticFormer backbone where ElasticSelector is integrated (larger dimensions for early layers capturing fine details, smaller for deeper semantic layers).

After obtaining the $\mathbf{F}_{\text{flat}}$, we employ a lightweight two-layer MLP, $f_{\text{pred}}$, to predict the foreground proportion $\hat{\rho}$ for ElasticSelector in each stage as shown in \Cref{fig:2}, marked as stage $t$:
\begin{equation}
    \hat{\rho^t} = \sigma(f_{\text{pred}}(\mathbf{F}^t_{\text{flat}}))
\end{equation}
where layer of Sigmoid ensures that $\hat{\rho}$ lies between 0 and 1. $\hat{\rho}$ will be used for top-k selection, shown in \Cref{eq:5}.

\begin{table*}
    \renewcommand{\arraystretch}{1.2}
    \centering
    \resizebox{\textwidth}{!}{
    \begin{tabular}{c|c|ccc|cccc}
    \hline
         Method & Backbone & GFLOPs-F & GFLOPs-B & GFLOPs-O & $\text{AP}_{50}$ & $\text{AP}_\text{small}$ & $\text{AP}_\text{meduim}$ & $\text{AP}_\text{large}$ \\
         \hline
         CascadeRCNN \cite{cai2018cascade} & ResNet-101 \cite{he2016deep} & 15.54 & 295.24 & 310.78 & - & 0.227 & 0.579 & 0.765\\
         ClusDet \cite{yang2019clustered} & ResNet-50 \cite{he2016deep} & 10.35 & 196.71 & 207.07 & 0.718 & 0.219 & 0.696 & 0.782\\
         DMNet \cite{li2020density} & ResNet-50 & 10.35 & 196.71 & 207.07 & 0.540 & 0.119 & 0.371 & 0.714\\
         FasterRCNN \cite{fan2022speed} & ResNet-50 & 10.35 & 196.71 & 207.07 & 0.705 & 0.203 & 0.712 & 0.760\\
         GigaDet \cite{chen2022towards} & CSP-DarkNet-53 & 4.61 & 87.59 & 92.20 & 0.684 & 0.210 & 0.599 & 0.762\\
         PAN \cite{fan2022speed} & ResNet-50 & 10.35 & 196.71 & 207.07 & 0.715 & 0.256 & 0.719 & 0.768\\
         RetinaNet \cite{lin2017focal} & ResNet-101 & 15.77 & 299.62 & 315.39 & - & 0.221 & 0.561 & 0.740\\
         \hline
         Dynamic-Head \cite{dai2021dynamic} & Swin-T \cite{Swin} & 5.74 & 109.1 & 114.8 & 0.592 & 0.165 & 0.537 & 0.694\\
         Dynamic-Head+SparseFormer & SparseFormer \cite{SparseFormer} & 6.29 & 58.35 & 64.64 & 0.771 & 0.364 & 0.740 & 0.863\\
         \rowcolor{OliveGreen!20} Dynamic-Head+ElasticFormer & ElasticFormer(Ours) & 6.91 & 6.58 & 13.49 & 0.782 & 0.409 & 0.762 & 0.860\\
         \hline
         DINO \cite{DINO} & ResNet-50 \cite{he2016deep} & 6.21 & 118.02 & 124.24 & 0.542 & 0.289 & 0.530 & 0.592\\
         \rowcolor{RoyalBlue!20} DINO+ElasticNet & ElasticNet(Ours) & 6.79 & 9.75 & 16.54 & 0.754 & 0.388 & 0.734 & 0.791\\
         \hline
         DINO \cite{DINO} & Swin-T \cite{Swin} & 6.64 & 126.19 & 132.84 & 0.606 & 0.367 & 0.612 & 0.649\\
         DINO+SparseFormer & SparseFormer \cite{SparseFormer} & 6.90 & 68.81 & 75.71 & 0.780 & 0.508 & 0.781 & 0.823\\
         \rowcolor{OliveGreen!20} DINO+ElasticFormer & ElasticFormer(Ours) & 7.02 & 6.11 & 13.13 & 0.806 & 0.515 & 0.802 & 0.849\\
         \hline
    \end{tabular}}
    \caption{\textbf{Computational cost and accuracy comparison of different models on PANDA.} GFLOPs are calculated exclusively for the backbone and neck when applicable. ``F'', ``B'' and ``O'' denote foreground, background and overall respectively.}
    \label{tab:1}
\end{table*}

\noindent\textbf{Loss function for ElasticSelector.}
As a novel module, ElasticSelector requires its own loss function for training. We propose a novel loss that accelerates its convergence while promoting balance between calculation and accuracy:
\begin{equation}
    \mathcal{L}_{\text{select}} = (\rho - \hat{\rho})^2 + \gamma \cdot \hat{\rho},
\end{equation}
where $\gamma$ lays in the interval $(0,1)$, motivating smaller $\hat{\rho}$ through $L_1$ penalty. We set $\gamma$ to 0.1 through experiments. 

Considering that bounding-box annotations are not always available—such as in weakly supervised object detection scenarios—we further extend the supervisory signal $\rho$ to support model-generated pseudo labels. Specifically, $\rho$ can be computed from either human-annotated ground truth (GT) or from pseudo bounding boxes (pseudo-GT) predicted by the model itself during training. This design allows ElasticSelector to perform label-free self-supervised optimization, making it applicable in both fully and weakly supervised settings.

The residual between $\rho$ and the prediction $\hat{\rho}$ drives ElasticSelector to learn overall image features and accurately infer foreground proportion; concurrently, penalizing the predicted value itself encourages ElasticSelector to minimize its magnitude, once accurate predictions are secured, thereby reducing computational overhead.

As for our ElasticFormer, we apply the loss function to each stage, enabling them to learn the characteristics of their respective feature maps, thereby further enhancing the adaptability of our model. The overall loss function is defined as follows:
\begin{equation}
    \mathcal{L}_{\text{total}} = \mathcal{L}_{\text{bbox}} + \mathcal{L}_{\text{select}}.
\end{equation}

$\mathcal{L}_{\text{bbox}}$ is the loss generally used for detection task, helping detector to locate targets precisely.

\noindent\textbf{3-Phase training strategy.}
When training ElasticSelector using pseudo-GT, the model initially generates highly disordered bounding boxes. Under these conditions, the computed density values cannot effectively guide the optimization of ElasticSelector’s parameters and may even exert a negative effect on learning. Moreover, since the backbone’s token budget is directly determined by the density predicted by ElasticSelector, underestimation of density leads to a reduction in supervisory signals for the backbone, thereby impairing its convergence.

We therefore seek to train ElasticSelector with more stable pseudo‐GT and to allow the backbone to utilize more tokens during the early period of training, thereby accelerating its convergence. However, a full decoupling of the two branches is not warranted, as achieving a balance between computational overhead and detection accuracy necessitates their coordinated operation. To this end, we introduce a 3‐phase training strategy:

First, we freeze ElasticSelector and force the backbone to train on all token, ignoring $\hat{\rho}$. This guarantees that the backbone receives sufficient supervisory signal and prevents useless updates to ElasticSelector.

Second, once the backbone’s outputs have stabilized, we unfreeze ElasticSelector to drive its initial convergence and obtain more reliable density estimates. However, $\hat{\rho}$ is still ignored in this phase.

Third, we put $\hat{\rho}$ in use, enabling the network to autonomously balance computational load against detection accuracy.

\section{Experiment}
\subsection{Datasets}
Our primary evaluation is conducted on the PANDA \cite{PANDA} dataset, the first human-centric gigapixel-level benchmark featuring 13 training and 5 testing scenes with approximately 16 million annotated bounding boxes. The average image resolution of this dataset, 25,000 $\times$ 16,000 pixels, presents unprecedented computational demands, establishing a rigorous testbed for efficiency assessment.

\subsection{Evaluation metrics}
Model efficiency is quantified using FLOPs and standard COCO metrics, with particular emphasis on $\text{AP}_{50}$, $\text{AP}_\text{small}$ ($<96\times 96$), $\text{AP}_\text{medium}$ ($96\times 96 \sim 288\times 288$) and $\text{AP}_\text{large}$ ($>288\times 288$). We further record the FLOPs on foreground (FLOPs-F) and background (FLOPs-B) to provide a more fine‑grained evaluation of model. To ensure fair comparison, all detectors process uniformly cropped 1280×800 windows from test images. Given the dynamic token selection mechanism in  ElasticFormer, its FLOPs are estimated by averaging measurements across the test set. All implementations utilized the MMDetection \cite{chen2019mmdetection} framework under standardized 36-epoch training protocols.

\subsection{Data analysis on PANDA}
As presented in \Cref{tab:1}, traditional detectors consistently exceed 200 GFLOPs for backbone computation on PANDA \cite{PANDA}, whereas modern Transformer-based approaches like Dynamic-head \cite{dai2021dynamic} and DINO \cite{DINO} reduces computational costs to sub-100 GFLOPs. SparseFormer achieves further reduction to 75.71 GFLOPs through fixed-ratio token selection while delivering significant AP gains.

Whether integrated with Dynamic‑Head \cite{dai2021dynamic} or DINO \cite{DINO}, our ElasticFormer delivers superior performance under significantly lower computational cost. Combined with DINO, our backbone outperforms the SparseFormer \cite{SparseFormer} by 3.3\% AP$_{50}$ while consuming only 13.13 GFLOPs. This represents a 82.7\% computational reduction compared to SparseFormer and an 90.2\% reduction relative to Swin-T \cite{Swin}, demonstrating highly efficient sparse computation. The computational reduction stems from dramatically curtailing redundant background computations: the ElasticFormer expends merely 8.9\% of the computational resources wasted by its SparseFormer counterpart on background regions.

\begin{table*}
    \renewcommand{\arraystretch}{1.2}
    \centering
    \resizebox{\textwidth}{!}{
    \begin{tabular}{c|c|ccc|c|c}
        \hline
        Method & Backbone & GFLOPs-F & GFLOPs-B & GFLOPs-O & mAP(\%) & CorLoc(\%)\\
        \hline
        WSOD$^2$ \cite{zeng2019wsod2}+GoogleNet & GoogleNet \cite{szegedy2015going} & 5.56 & 12.38 & 17.94 & 37.9 & 52.8\\
        WSOD$^2$+VGG16 & VGG16 \cite{simonyan2014very} & 36.88 & 86.05 & 122.93 & 47.2 & 59.5\\
        WSOD$^2$+ResNet-50 & ResNet-50 \cite{he2016deep} & 63.01 & 111.90 & 174.91 & 46.7 & 59.9\\
        WSOD$^2$+ResNet-101 & ResNet-101 \cite{he2016deep} & 70.83 & 165.27 & 236.10 & 48.9 & 61.1\\
        \hline
        Modified-DINO+Swin-T & Swin-T \cite{Swin} & 11.49 & 26.80 & 38.29 & 49.9 & 63.7\\
        Modified-DINO+SparseFormer & SparseFormer \cite{SparseFormer} & 9.11 & 13.01 & 22.12 & 50.3 & 63.1\\
        \rowcolor{OliveGreen!20} Modified-DINO+ElasticFormer(Ours) & ElasticFormer(Ours) & 7.45 & 3.26 & 10.71 & 50.9 & 63.9\\
        \hline
    \end{tabular}
    }
    \caption{\textbf{Comparisons of FLOPs, mean average precision and correct localization on different backbones in WSOD tasks.} We modified DINO’s loss function using a Multiple Instance Learning (MIL) based loss to make it suitable for the WSOD task.}
    \label{tab:2}
\end{table*}

\begin{figure}
    \centering
    \includegraphics[width=1\linewidth]{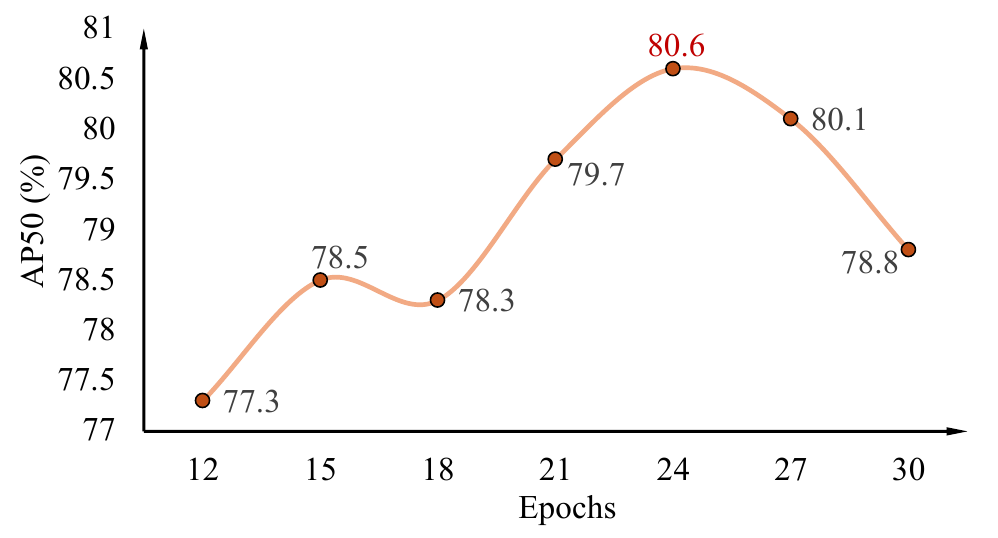}
    \caption{\textbf{AP$_{50}$ (\%) vs. the length of Phase 1.} As Phase 1 becomes longer, AP$_{50}$ first increases and then decreases with the maximum at 24 epochs.}
    \label{fig:4}
\end{figure}

\subsection{3-Phase Training Strategy}

\noindent\textbf{Comparison of strategies.}
As shown in \Cref{tab:3}, 3-phase strategy is consistently better than 1-phase (utilizing ElasticSelector all the time). The results demonstrate that the 3-phase strategy effectively mitigates the impact of early $\hat{\rho}$ noise on overall training. Moreover, with 3-phase strategy, results on GT under-performs pesudo-GT. This is because the noise in pseudo-GT can effectively enhance model generalization during joint optimization (Phase 3).

\noindent\textbf{Balancing Phase 1 and 3.}
To mitigate mutual interference between the ElasticSelector pathway and the main backbone during early training, we implement a controlled 3-phase training regimen (detailed in Section 3.4) that gradually activates end-to-end coupling. The initial transition point is set after epoch 12, where ElasticSelector begins training without yet influencing the backbone. Analysis of loss convergence patterns during 1-phase training determines that Phase 2 spans 3 epochs, sufficient for ElasticSelector stabilization before initiating joint optimization (Phase 3).

As Phase 1 and 3 share the total epochs of 33, the key of optimizing 3-phase strategy is to balance them. \Cref{fig:4} demonstrates critical tradeoffs: the shorter Phase 1 cause the noise in early-training, the shorter Phase 3 hinder the joint optimization. They both decrease the final accuracy.

\noindent\textbf{Discussion on lower selection rates of stage3.}
During the ablation experiments on training strategies, we observed a phenomenon in \Cref{tab:2}: the sparsity rate at stage 3 is noticeably lower than that of other stages. Examining the model structure, we speculate that this may reflect an adaptation of Elastic Selection to network depth. As shown in \Cref{fig:2}, the number of Elastic Selection executions corresponds to the depth, and a lower sparsity rate at greater depth may imply selecting fewer but more critical windows each time, thereby achieving foreground coverage through multiple selections. As this is not a central focus of our work, only a tentative discussion is provided here.

\begin{table}
    \renewcommand{\arraystretch}{1.2}
    \centering
    \resizebox{\columnwidth}{!}{
    \begin{tabular}{c|c|c|c|c|c|c}
        \hline
        \multirow{2}{*}{Strategies} & \multirow{2}{*}{Supervision} & \multicolumn{4}{c|}{Windows selection rates (\%)} & \multirow{2}{*}{AP$_{50}$}\\
        \cline{3-6}
         &  & Stage1 & Stage2 & Stage3 & Stage4 & \\
        \hline
        \multirow{2}{*}{1-phase} & GT & 15.11 & 15.32 & 12.79 & 15.73 & 0.791 \\
        & Pseudo-GT & 14.02 & 14.24 & 9.80 & 12.97 & 0.784 \\
        \hline
        \multirow{2}{*}{3-phase} & GT & 14.15 & 14.55 & 13.28 & 14.95 & 0.798 \\
        & Pseudo-GT & \textbf{11.31} & \textbf{10.41} & \textbf{7.56} & \textbf{12.09} & \textbf{0.806} \\
        \hline
    \end{tabular}
    }
    \caption{\textbf{Windows selection rates and AP$_{50}$ across different configures and stages.} Pseudo-GT means outputs from the detection head. Rates on Stage 3 is always the lowest.}
    \label{tab:3}
\end{table}

\subsection{Backbone-agnostic Study}

We keenly observe that the proposed ElasticSelector is backbone‑agnostic and can be applied to both convolution‑based and Transformer‑based architectures. After implementing ElasticFormer on Transformer, we further obtained ElasticNet by replacing the self‑attention modules with convolutional layers. As shown in \Cref{tab:1}, when combined with DINO and compared against ResNet‑50, this validation backbone achieves a 87\% reduction in computation while delivering competitive detection quality. This provides strong evidence that ElasticSelector can be applied to various backbones, dynamically adjusting computation according to the foreground proportion.

\subsection{WSOD Study}
\textbf{Methodology.}
Since bounding box annotations are absent in Weakly Supervised Object Detection (WSOD), the loss in the DINO \cite{DINO} detection head based on them must be replaced. Inspired by CDTR \cite{chen2025clip} and other Transformer-based approaches, we incorporate the Smooth L1 loss, training with numerous pseudo BBoxes generated by the model.

\noindent\textbf{Datasets.}
We conduct weakly supervised experiments on the PASCAL VOC 2007 \cite{everingham2010pascal} dataset, which is split into training, validation, and test subsets. It comprises 9,963 annotated images containing 24,640 object instances spanning 20 distinct categories. The images have an overall resolution on the order of $10^5$ pixels, placing them in the non-HRW shots regime. This configuration provides a robust benchmark for evaluating ElasticFormer’s performance across diverse detection scenarios and target classes.

\noindent\textbf{Evaluation metrics.}
Moving to evaluation, we use mAP and CorLoc as the metrics and the computational costs are measured on 224×224 resolution inputs. To ensure fair comparison, all transformer-based and convolution-based detectors are evaluated after the same epochs of training separately. We estimate the computation on ElasticFormer by averaging measurements across the test set due to elastic computation.

\noindent\textbf{Data analysis.}
As evidenced by \Cref{tab:3}, our sparsification approach reduces computational load by up to 70\% on non-HRW shots, compared with Swin-T \cite{Swin}. Further, ElasticFormer achieves gains in both mAP and CorLoc. It is noteworthy that in WSOD tasks, computation devoted to foreground regions increases markedly on non-HRW shots. This demonstrates that ElasticFormer can dynamically adjust its resource allocation to more effectively extract features in object-relevant areas.

This significant efficiency gain persists even when target density falls below typical HRW thresholds, thereby demonstrating the generalizability beyond specialized high-resolution contexts, including WSOD tasks, of our method.

\begin{figure}[h]
    \centering
    \includegraphics[width=1\linewidth]{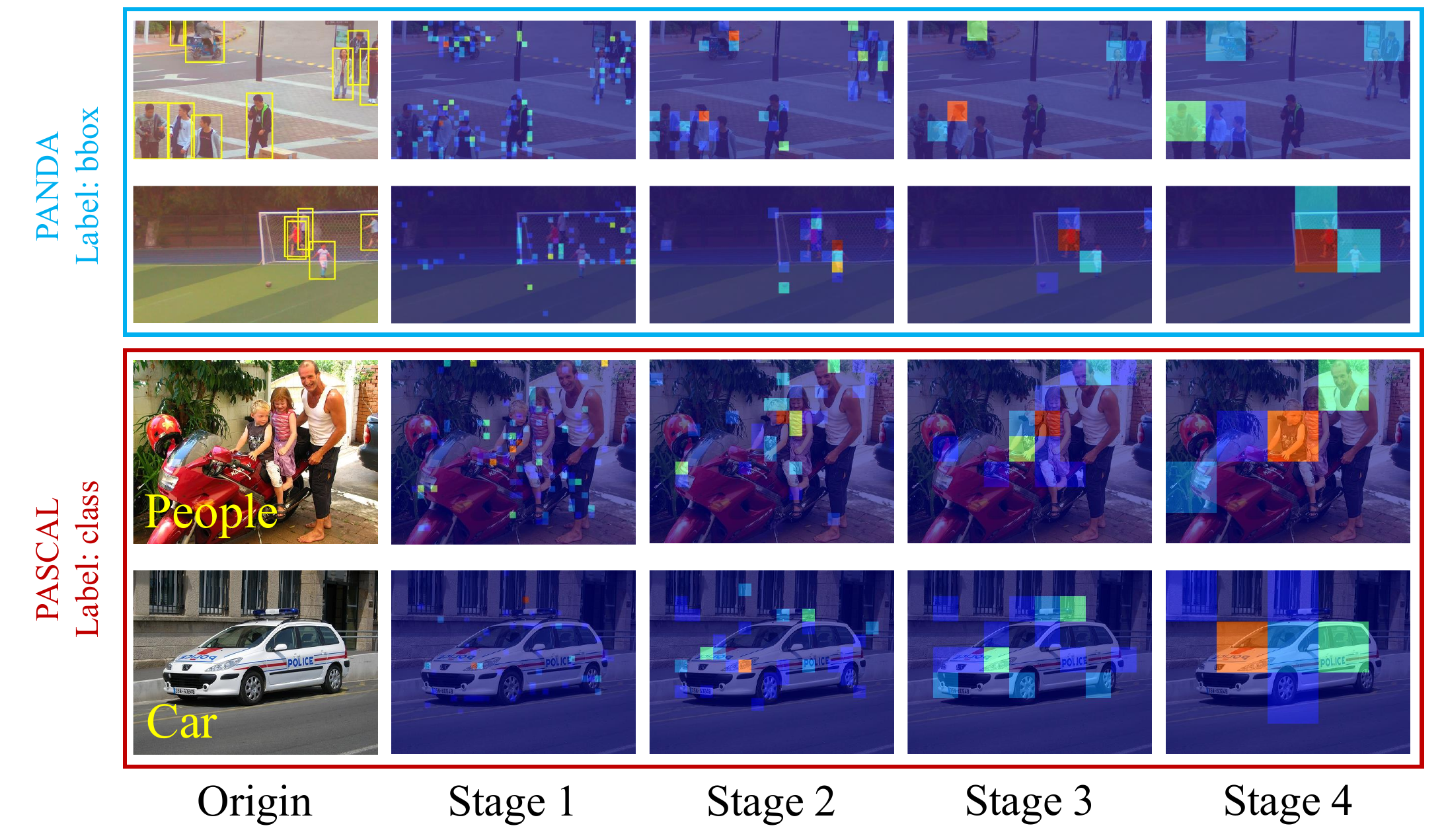}
    \caption{\textbf{Visualization of scored windows.} We illustrate scored windows from each stage of ElasticFormer, with the top two rows from PANDA \cite{PANDA} and the others from PASCAL VOC 2007 \cite{everingham2010pascal} for WSOD task. Highlighted windows indicate regions requiring
    further feature extraction.}
    \label{fig:6}
\end{figure}

\subsection{Visualization of Sparse Selection}

\Cref{fig:6} illustrates our window scoring and sparsification mechanism with a thermal color map: high-scoring windows are shown in red, and low-scoring regions in blue. By using ElasticSelector’s score-driven sparse selection, we retain only $\hat{\rho} \cdot n$ windows for detailed feature extraction. This approach markedly cuts computation on uninformative backgrounds and on areas with subtle features.

This efficiency gain arises from our target-coverage-ratio–based elastic computation modulation, which enables ElasticFormer to achieve high detection efficiency. \Cref{fig:6} also includes examples from both the PANDA \cite{PANDA} and PASCAL VOC 2007 \cite{everingham2010pascal} datasets, demonstrating the method’s effectiveness across different detection scenarios, tasks and object classes.

\subsection{Ablation Study}

\noindent\textbf{Ablation study on $\alpha$ and $\gamma$.} We conduct a joint ablation study by selecting three values for each. Here we introduce a new metric, F-ratio, $\text{F-ratio}=(\text{FLOPs-F}/\text{FLOPs-O})\times 100\%$. This metric reflects the backbone’s attention to foreground regions. Increasing $\gamma$ reduces computational cost at the expense of AP50, highlighting the trade-off between efficiency and accuracy. F-ratio shows no significant association with $\gamma$.

In contrast, decreasing $\alpha$ proportionally reduces the supervised signal $\rho$, which also lowers FLOPs-O. Interestingly, AP$_{50}$ does not follow the same trend; it reaches its peak when $\alpha = 0.8$. By examining F-ratio, we observe that smaller $\alpha$ values correspond to higher F-ratios, indicating that the backbone focuses more on foreground regions. Therefore, tuning $\alpha$ enables a trade-off between computational cost and foreground attention, leading to optimal detection performance.

\begin{table}[ht]
    \renewcommand{\arraystretch}{1.2}
    \resizebox{\columnwidth}{!}{
    \centering
        \begin{tabular}{c|c|c|c|c|c}
        \hline
        $\alpha$ & $\gamma$ & AP$_{50}$ & GFLOPs-O & GFLOPs-F & F-ratio (\%) \\
        \hline
        \multirow{3}{*}{1.0} & 0.1 & \underline{0.802} & 14.98 & 6.84 & 45.7\\
        & 0.2 & 0.799 & 13.17 & 6.00 & 45.6 \\
        & 0.3 & 0.793 & 11.44 & \textbf{5.25} & 45.9\\
        \hline
        \multirow{3}{*}{0.8} & 0.1 & \textbf{0.806} & 13.13 & 7.02 & 53.5 \\
        & 0.2 & 0.801 & 11.91 & 6.32 & 53.1 \\
        & 0.3 & 0.795 & \underline{10.28} & 5.48 & 53.3 \\
        \hline
        \multirow{3}{*}{0.6} & 0.1 & 0.799 & 11.52 & 6.36 & \textbf{55.2} \\
        & 0.2 & 0.791 & 10.38 & 5.68 & 54.7\\
        & 0.3 & 0.794 & \textbf{9.77} & \underline{5.38} & \underline{55.1} \\
        \hline
        \end{tabular}
    }
    \caption{\textbf{AP$_{50}$ and FLOPs across different $\alpha$ and $\gamma$.} F-ratio is the proportion of foreground computation to the total, indicating a backbone’s attention to foreground regions.}
    \label{tab:4}
\end{table}

\noindent\textbf{Ablation study on scoring mechanism.} We further conduct ablation experiments on different window scoring mechanisms, including small learned objectness and gradient density. The results show that small learned objectness can slightly improve foreground selection accuracy, but it requires an additional prediction head and thus introduces noticeable latency. Gradient density captures edge and texture information, which is sensitive to noise and incurs extra computational overhead. In contrast, our variance-based scoring mechanism is lightweight and does not rely on additional network components, leading to significantly lower latency while maintaining competitive detection accuracy. This demonstrates a superior balance between efficiency and accuracy.

\begin{table}[h]
    \renewcommand{\arraystretch}{1.2}
    \resizebox{\columnwidth}{!}{
    \centering
        \begin{tabular}{c|c|c|c|c}
        \hline
        Scoring Mechanism & AP$_{50}$ & GFLOPs-F & F-ratio (\%) & Latency (ms) \\
        \hline
        Variance-based(Ours) & \textbf{0.806} & 7.02 & 53.5 & \textbf{0.397} \\
        Small objectness & 0.803 & \textbf{7.14} & \textbf{53.8} & 1.213 \\
        Gradient density & 0.798 & 5.74 & 41.7 & 0.829 \\
        \hline
        \end{tabular}
    }
    \caption{\textbf{AP$_{50}$, FLOPs and latency across different scoring machanism.} Latency is measured as the average over four stages, 100 forward passes of the backbone.}
    \label{tab:5}
\end{table}

\noindent\textbf{Ablation study on adaptive sparsity rates.} SparseFormer \cite{SparseFormer} also adopts the concept of sparsity rate, but it uses a fixed parameter preset to 0.7. We choose this parameter from $[0.1, 0.3, 0.5, 0.7, 1.0]$ to study SparseFormer’s performance under different sparsity rates. The experiments are conducted with both DINO \cite{DINO} and Dynamic-Head \cite{dai2021dynamic}.

\begin{figure}[h]
    \centering
    \includegraphics[width=1\linewidth]{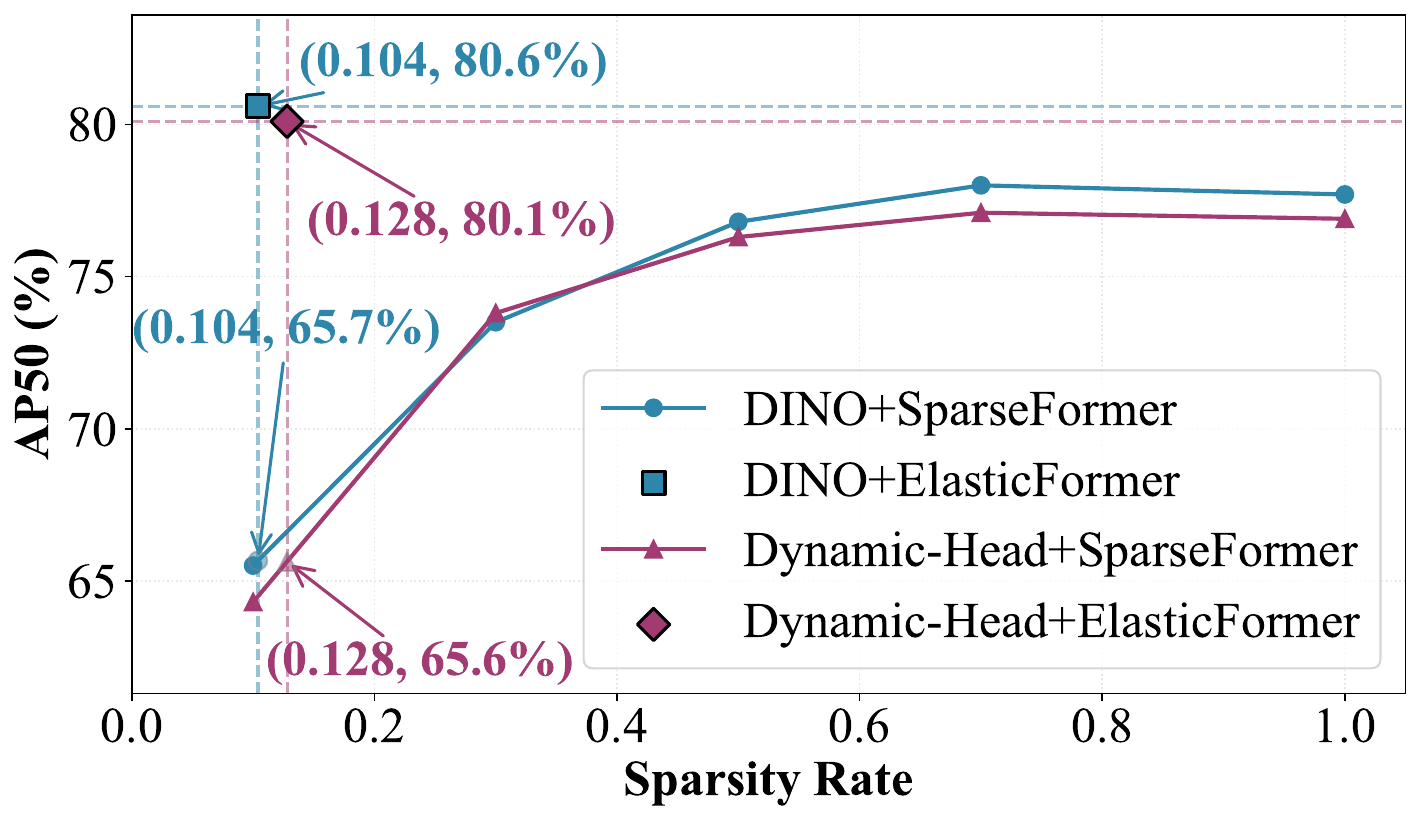}
    \caption{\textbf{AP$_{50}$ vs. sparsity rate on SparseFormer and ElasticFormer.} The annotated data in the figure is obtained through linear interpolation.}
    \label{fig:7}
\end{figure}

Regardless of the detection head, SparseFormer suffers a sharp drop in AP$_{50}$ when the sparsity rate is small. With the same sparsity rate as ElasticFormer, its AP$_{50}$ is consistently at least 15\% lower. This highlights the importance of dynamic sparsity in addressing target number variations in HRW scenarios.

\section{Conclusion}
We propose ElasticFormer, a Vision Transformer that dynamically allocates computation for HRW shot detection via its lightweight ElasticSelector module. By estimating target-coverage ratios $\hat{\rho}$ and prioritizing high-variance regions, it reduces FLOPs by 90.2\% versus Swin-T \cite{Swin} and 82.7\% versus SparseFormer \cite{SparseFormer} while achieving 80.6\% AP50 on PANDA \cite{PANDA}. The framework’s backbone-agnostic design enables adaptive token utilization across stages, cutting background computation by 91.1\%. This elastic paradigm establishes a new standard for efficient large-scale vision systems, particularly for real-time applications with constrained resources.

\newpage

\section*{Acknowledgments}
This work was supported by National Science and Technology Major Project (No. 2022ZD0119402), 
National Natural Science Foundation of China (No. 6252510), 
“Pioneer” and “Leading Goose” R\&D Program of Zhejiang (No. 2024C01142), 
and Tsinghua University – Migu Xinkong Culture Technology (Xiamen) Co., Ltd. Joint Research Center for Intelligent Light Field and Interaction Technology, Tsinghua University - Joint R\&D Project (R2511DKW) – Phase 2. We also acknowledge the KLATASDS-MOE.

{
    \small
    \bibliographystyle{ieeenat_fullname}
    \bibliography{main}
}


\end{document}


\maketitle

\section{Details of Training and Inference}

\textbf{Training configs.}
We train ElasticFormer with 36 epochs on RTX 4090 GPUs, using learning rate $l=2e\text{-}4$. We use AdamW as the optimizer, linearly warming up the learning rate to its preset value over the first two epochs, and employ a MultiStepLR schedule to reduce it by a factor of ten at the 18th and 30th epochs.

\begin{figure}[htbp]
    \centering
    \subfloat[1-Phase Strategy]{%
		\includegraphics[width=1\linewidth]{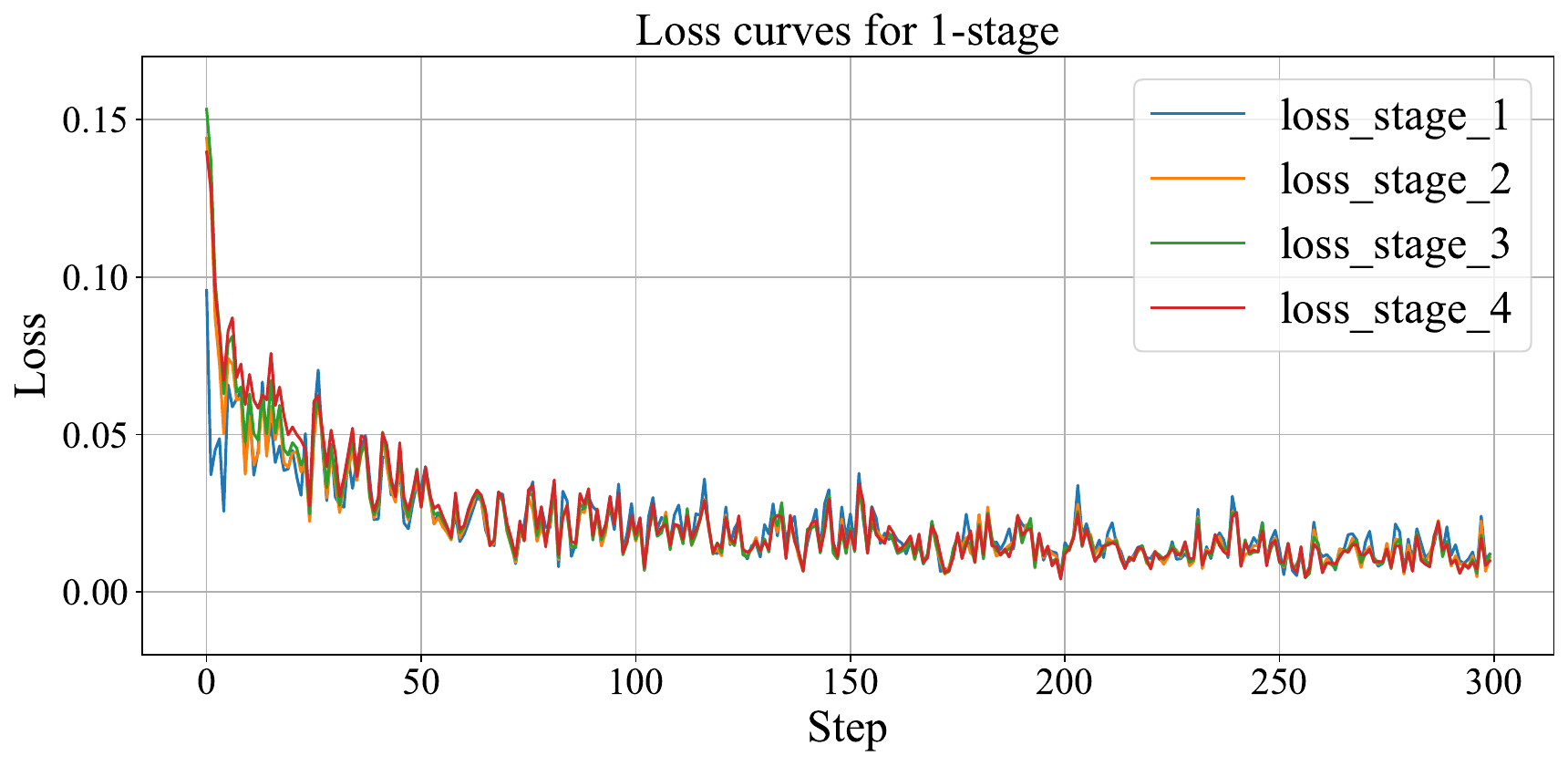}}\\
    \subfloat[3-Phase Strategy]{%
		\includegraphics[width=1\linewidth]{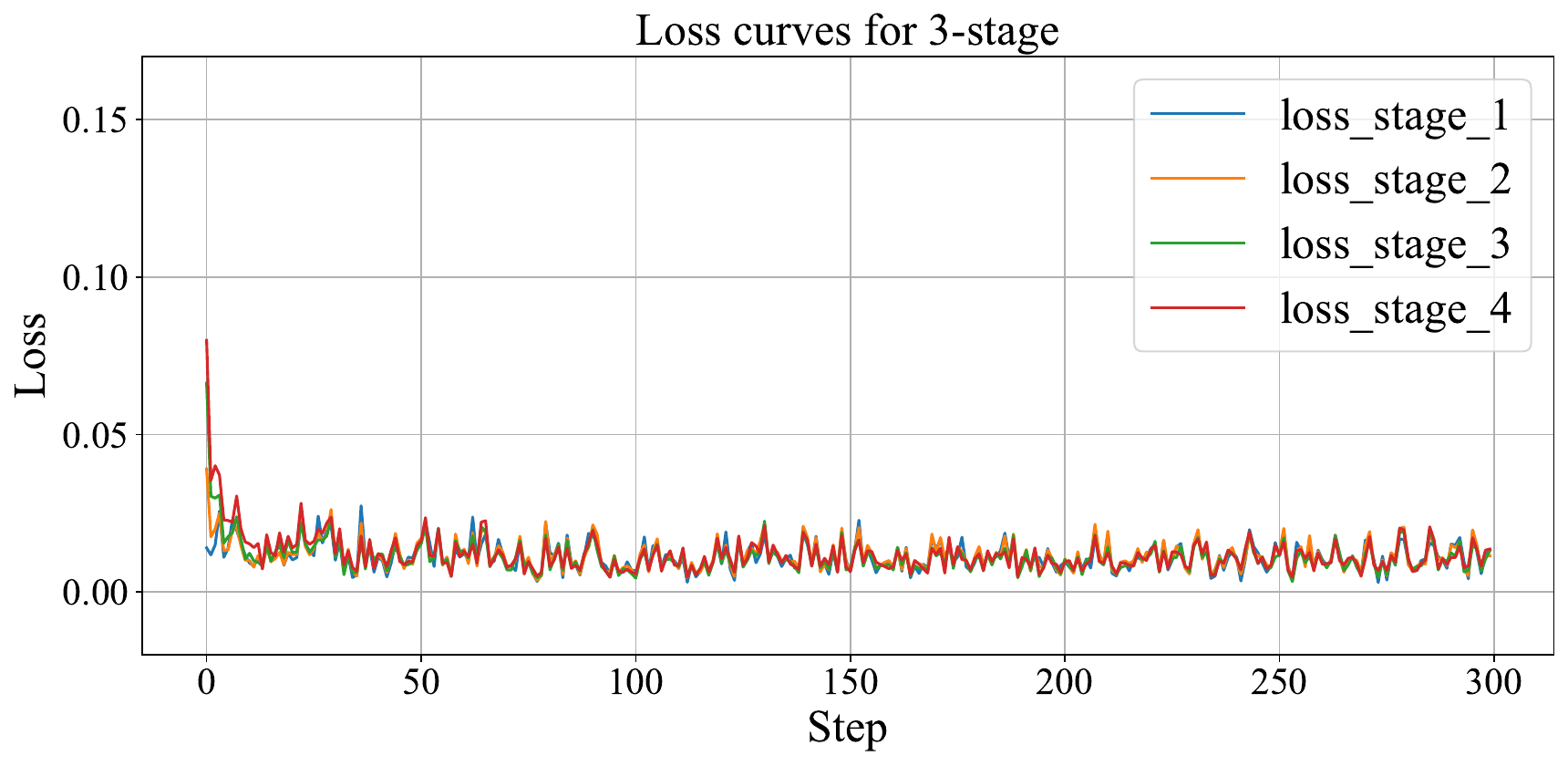}}
    \caption{\textbf{Comparison of the loss for different strategies.} We demonstrate 4 loss curves in each sub-figure, representing the loss for each stage in ElasticFormer.}
    \label{Figure 1}
\end{figure}

\vspace{2mm}
\noindent\textbf{Loss for 3-phase training.}
We compare the loss progression across the four training stages before and after adopting the 3-phase strategy, illustrating the results in \Cref{Figure 1}.

It is worth noting that all loss values were logged from the start of ElasticSelector training, meaning the curves for the two strategies are aligned only on a relative scale rather than a common absolute scale. Nevertheless, the convergence behavior matches that shown in the figure: the ElasticSelector module trained with the 3-phase strategy is more stable and achieves faster convergence in relative terms compared to when the strategy is omitted.

\vspace{2mm}
\noindent\textbf{Inference configs.}
During inference, we divided large-resolution images into patches of $6000\times 3600$ with an overlap of 600 pixels between patches for the PANDA \cite{PANDA} dataset, resizing images into $512\times 512$ for PASCAL VOC 2007 \cite{everingham2010pascal}.

\clearpage
\onecolumn
\section{Visualization Results}

\begin{figure}[htbp]
    \centering
    \includegraphics[width=1\linewidth]{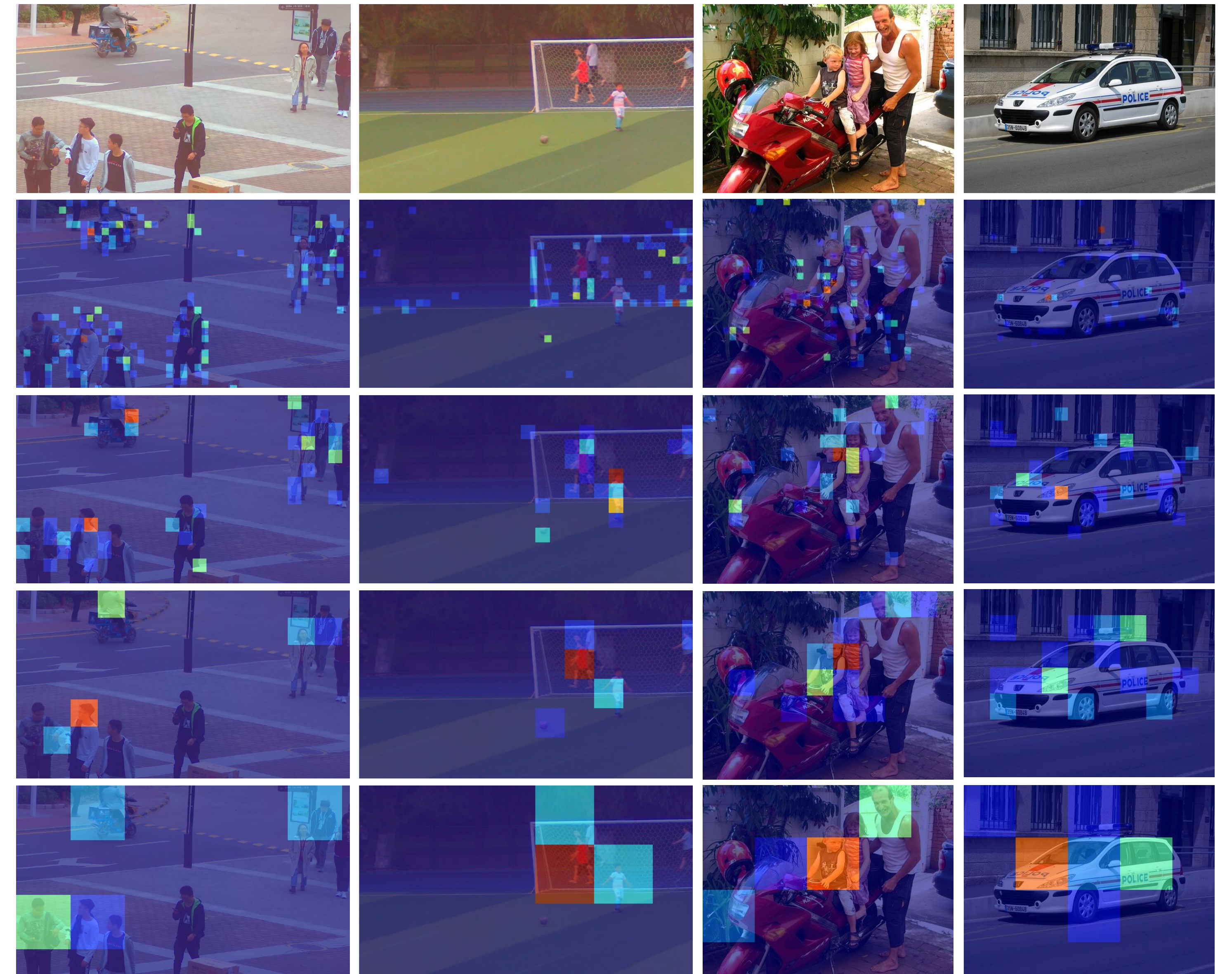}
    \caption{\textbf{Visualization of scored windows.} Left two columns come from PANDA \cite{PANDA} and right two columns come from PASCAL VOC 2007 \cite{everingham2010pascal}.}
    \label{Figure 2}
\end{figure}

To more clearly demonstrate the differences between the attention regions at each stage, we enlarge and reproduce the paper’s visualizations in \Cref{Figure 2} to facilitate direct comparison.
As shown in \Cref{Figure 2}, each stage corresponds to a different window size within the pyramid structure: the later stages (depicted lower in the image) employ larger windows that encompass broader regions.

\clearpage
\section{Code in PyTorch}
The public implementation is available at \url{https://github.com/Purecoo00/ElasticFormer}.
Algorithms~\ref{alg:select} and~\ref{alg:calculate} show the selection methods of \texttt{ElasticStage} and \texttt{ElasticFormer}, and the foreground-target helpers. Window scoring, attention, and detector integration are omitted.

ElasticSelector retains one predicted ratio per image using \texttt{squeeze(-1)}. The first two training phases use all windows; the third phase and inference select $\lceil\hat{\rho}N_w\rceil$ windows per image, retaining at least one. The squared error and ratio penalty are computed per image before averaging over the batch and four stages. The selector is frozen in the first phase.

Foreground targets use clipped bounding-box union area and $\alpha=0.8$ (Equation~(1)). Boxes and dimensions share the augmented image coordinates (\texttt{img\_shape}). Pseudo boxes are predicted under \texttt{torch.no\_grad()} with \texttt{rescale=False} and a confidence threshold of 0.3; GT boxes require no score filtering.

\definecolor{codeblue}{rgb}{0.25,0.5,0.5}
\definecolor{codekw}{rgb}{0.85,0.18,0.50}
\lstset{
  backgroundcolor=\color{white},
  basicstyle=\footnotesize\ttfamily,
  columns=fullflexible,
  breaklines=true,
  showstringspaces=false,
  captionpos=b,
  commentstyle=\color{codeblue},
  keywordstyle=\color{codekw},
  upquote=true,
  keepspaces=true
}

\begin{algorithm}[H]
\caption{Per-image window selection and selection loss}
\label{alg:select}
\begin{lstlisting}[language=python,firstnumber=1,firstline=10,lastline=31]
def select_windows(self, scores, ratios):
    selections = []
    for image in range(scores.shape[0]):
        # Keep each image's budget independent, including batches above one.
        if self.training and not self.use_dynamic_topk:
            count = scores.shape[1]
        else:
            count = min(
                scores.shape[1],
                max(1, math.ceil(float(ratios[image].detach()) * scores.shape[1])),
            )
        selections.append(scores[image].topk(count).indices)
    return selections


def selection_loss(self, targets):
    return torch.stack(
        [
            ((stage.last_ratios - targets) ** 2 + self.select_gamma * stage.last_ratios).mean()
            for stage in self.stages
        ]
    ).mean()
\end{lstlisting}
\end{algorithm}

\begin{algorithm}[H]
\caption{Foreground targets from pseudo boxes or GT (Equation (1))}
\label{alg:calculate}
\begin{lstlisting}[language=python,firstnumber=1,firstline=34,lastline=57]
def foreground_ratio(boxes, image_shape):
    """Measure clipped box coverage, counting overlapping areas only once."""
    height, width = image_shape[:2]
    if height <= 0 or width <= 0:
        raise ValueError("Image dimensions must be positive")
    if hasattr(boxes, "tensor"):
        boxes = boxes.tensor
    rectangles = []
    for x1, y1, x2, y2 in boxes.detach().cpu().tolist():
        x1, x2 = max(0.0, x1), min(float(width), x2)
        y1, y2 = max(0.0, y1), min(float(height), y2)
        if x2 > x1 and y2 > y1:
            rectangles.append(box(x1, y1, x2, y2))
    return unary_union(rectangles).area / (height * width) if rectangles else 0.0


def foreground_targets(instances, image_shapes, alpha=0.8, score_threshold=None):
    targets = []
    for result, shape in zip(instances, image_shapes):
        boxes = result.bboxes
        if score_threshold is not None:
            boxes = boxes[result.scores > score_threshold]
        targets.append(alpha * foreground_ratio(boxes, shape))
    return targets
\end{lstlisting}
\end{algorithm}

\clearpage
\bibliographystyle{ieeenat_fullname}
\bibliography{sup}